\documentclass[11pt,a4paper]{article}
\usepackage{times,latexsym}
\usepackage{url}
\usepackage[T1]{fontenc}
\usepackage{tikz}
\usetikzlibrary{fit}
\usetikzlibrary{tikzmark}
\usepackage[acceptedWithA]{tacl2021v1}

\usepackage{xspace,mfirstuc,tabulary}

\newif\iftaclinstructions
\taclinstructionsfalse 
\iftaclinstructions
\renewcommand{\confidential}{}
\renewcommand{\anonsubtext}{(No author info supplied here, for consistency with
TACL-submission anonymization requirements)}
\newcommand{\instr}
\fi

\iftaclpubformat 

\else

\fi

\pdfoutput=1

\usepackage{wrapfig}
\usepackage{times}
\usepackage{latexsym}
\usepackage{scrextend}
\usepackage{graphicx}
\usepackage{amsmath}
\usepackage{lipsum}  
\usepackage{arydshln}

\usepackage{amsfonts}
\usepackage[T1]{fontenc}

\usepackage[utf8]{inputenc}

\usepackage{graphicx}
\usepackage{dsfont}
\usepackage{bbold}
\usepackage{booktabs}

\usepackage{times}
\usepackage{latexsym}
\usepackage{multirow}
\usepackage{natbib}
\usepackage{amssymb}

\usepackage[T1]{fontenc}
\usepackage{tikz}
\usepackage{graphicx}
\usepackage{svg}

\usepackage[export]{adjustbox}

\usepackage{microtype}
\usepackage{hyperref}
\input{pyg_ramble}

\title{Retrieval-Pretrained Transformer: Long-range Language Modeling with Self-retrieval}

\newcommand\invisible[1]{}

\newcommand\new[1]{#1}

\author{Ohad Rubin ~~~~~~~ 
Jonathan Berant \\
The Blavatnik School of Computer Science, Tel Aviv University \\
 \small{\texttt{\{ohad.rubin,joberant\}@cs.tau.ac.il}}
}

\begin{document}
\maketitle
\begin{abstract}
Retrieval-augmented language models (LMs) have received much attention recently. However, typically the retriever is not trained jointly as a native component of the LM, but added post-hoc to an already-pretrained LM, which limits the ability of the LM and the retriever to adapt to one another. 
In this work, we propose the \emph{Retrieval-Pretrained Transformer} (RPT), an architecture and training procedure for jointly training a retrieval-augmented LM from scratch and apply it to the task of modeling long texts. 
Given a recently generated text chunk in a long document, the LM computes query representations, which are then used to retrieve earlier chunks in the document, located potentially tens of thousands of tokens before. Information from retrieved chunks is fused into the LM representations to predict the next target chunk. We train the retriever component with a semantic objective, where the goal is to retrieve chunks that increase the probability of the next chunk, according to a reference LM. We evaluate RPT on four long-range language modeling tasks, spanning books, code, and mathematical writing, and demonstrate that RPT improves retrieval quality and subsequently perplexity across the board compared to strong baselines. 
\end{abstract}
\begin{figure}[t]
    \centering
    \includegraphics[width=0.46\textwidth]{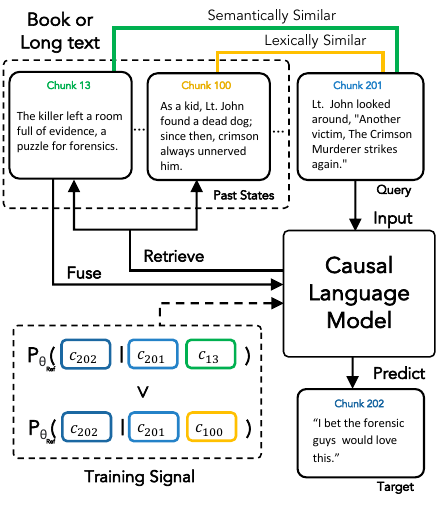}
    \caption{Retrieval-Pretrained Transformer (RPT) is a language trained from scratch with a native retrieval ability that can be applied to long texts (e.g., books). RPT takes a chunk of text as input, retrieves semantically-relevant chunks from the past to better predict the next chunk, and fuses these retrieved chunks into its representations. On top of a standard LM loss, the retriever is trained to retrieve chunks that increase the probability of the next chunk according to a \emph{reference LM}.
    }
    \label{fig:crime_scene}
\end{figure}
\;
\invisible{asd}
\section{Introduction}

Large language models (LMs) have had immense success recently \citep{gpt3,palm,opt,llama},
becoming a useful tool across disciplines. However, their success comes at a computational cost, due to increasing parameter counts for storing world knowledge \cite{switch} 
and growing context lengths that enable access to distant information, but incur a quadratic complexity penalty. 
Retrieval-augmented language modeling (RALM) alleviates this cost \citep{knnlm,spalm,retro,ram2023incontext},
as precise retrieval of relevant information can reduce memory and computation requirements. Moreover, RALM is beneficial for factuality, freshness and generalization without necessitating retraining, simply by swapping the retrieval index \cite{realm,rag,knnadapter}.

However, past work on RALM has by and large \emph{not} trained the retriever as a first-class component of the LM.
In some cases \citep{knnlm,spalm,retro}, the retriever was used only at test time, or remained fixed throughout training, preventing it from adapting to the LM generator. 
In other cases, the retriever component was jointly trained but only after a separate pretraining phase for both the retriever and LM
\citep{emdr2,atlas,reatt,unlimiformer}. Thus, the retriever was not pre-trained from scratch with the LM, and only a fraction of the training budget was allocated for joint training.

Recently, \newcite{trime} presented a retrieval-augmented LM that trains a retriever from scratch jointly with the LM, but (a) the retriever was trained to exploit \emph{lexical} information only, and (b) the retrieved information was 
not fused at the \emph{representation level} back into the LM.

In this work, we present the \textit{Retrieval-Pretrained Transformer (RPT)}, a  retrieval-augmented LM, where the retriever is a first-class component, trained jointly from scratch with the LM. 
RPT relies on two technical contributions. First, on the architecture side (see Fig.~\ref{fig:crime_scene}),
input representations for the retriever are computed from the LM representations themselves (a concept we dub \emph{self-retrieval}), and retrieved representations are fused back into the LM decoder for making next word predictions.
Second, we train the retriever with an \emph{auxiliary loss function} that encourages retrieving text fragments that increase the probability of generating the subsequent text. Specifically, given a recently-generated chunk $c_t$, the retriever is trained to retrieve chunks $c_i$ that increase the probability of $p_\text{scoring}(c_{t+1} \mid c_i, c_t)$ according to a reference \emph{scoring LM}. Fig.~\ref{fig:crime_scene} provides an illustrative example for a case where a crime scene is described, and a scoring LM shows the benefit of retrieving a chunk thousands of tokens away (chunk 13) compared to lexical retrieval, which leads to a chunk that is only superficially related (chunk 100). 
\new{Unlike existing retrieval-augmented models that use an auxiliary encoder for retrieval \cite{fid2,atlas,emdr2}, RPT is able to leverage its internal hidden states for retrieval after a \textit{single} pre-training stage, greatly simplifying joint training.
}

\new{We apply RPT to the problem of modeling long documents, such as books, articles and code, as those are naturally occurring examples of long-form content, where the entire index can be held within memory in a forward-pass. \\
We evaluate RPT on four language modeling tasks and find that it improves perplexity across all tasks, outperforming prior work \cite{blockrecurrent,memorizing} as well as strong baselines \cite{retro,trime}. Moreover, we show that RPT retrieves high-quality chunks compared to retrievers that rely on lexical information.
Based on our empirical findings, we argue RPT can pave the way toward a next generation of pre-trained LMs, where large corpora are used during pre-training, resulting in a language models where retrieval is a strongly embedded component. Our code is publicly available at \url{https://github.com/OhadRubin/RPT}}.

\section{Background}
To situate our contribution, we review relevant recent RALM work. We extend this to more related work in \S\ref{sec:related_work}.

Early work on RALMs, such as kNN-LM \cite{knnlm} used retrieval to improve language modeling by interpolating the next-word distribution produced by the LM with a distribution proposed through a \emph{test-time-only} retrieval mechanism. \newcite{retro} later proposed Chunked Cross-Attention (CCA), where retrieval is performed also at training time, and retrieval results are deeply fused into the representations produced by a Transformer decoder through attention. However, the retriever was trained separately and kept fixed during training, which prevented it from adapting to the LM over the course of training.

TRIME \cite{trime}, like this work, trained a retrieval-augmented LM from scratch where the retriever component and the decoder LM are trained jointly. 
Our work differs from TRIME in two aspects: First, TRIME, like kNN-LM, incorporates information from the retriever in a shallow manner through distribution interpolation, while we adopt CCA as a deeper fusion mechanism. Second, TRIME takes advantage of lexical clues for supervising the retriever, that is, given a query, the TRIME retriever learns to retrieve contexts that will lead to generating the same token as the query. We, on the other hand, use a scoring LM to evaluate what text chunks are relevant for increasing the probability of the chunk being generated, which leads to more semantic retrieval. This is similar to EPR \cite{epr}, which used this idea for learning to retrieve prompts for in-context learning, and perplexity distillation in Atlas \cite{atlas}. However, Atlas
does not train the retriever and LM from scratch and is an encoder-decoder model, more suitable for knowledge-intensive tasks. We, conversely, train from scratch and use a decoder model, more suitable for modeling long texts.

\begin{figure*}[h]
    \centering
    \includegraphics[width=1\textwidth]{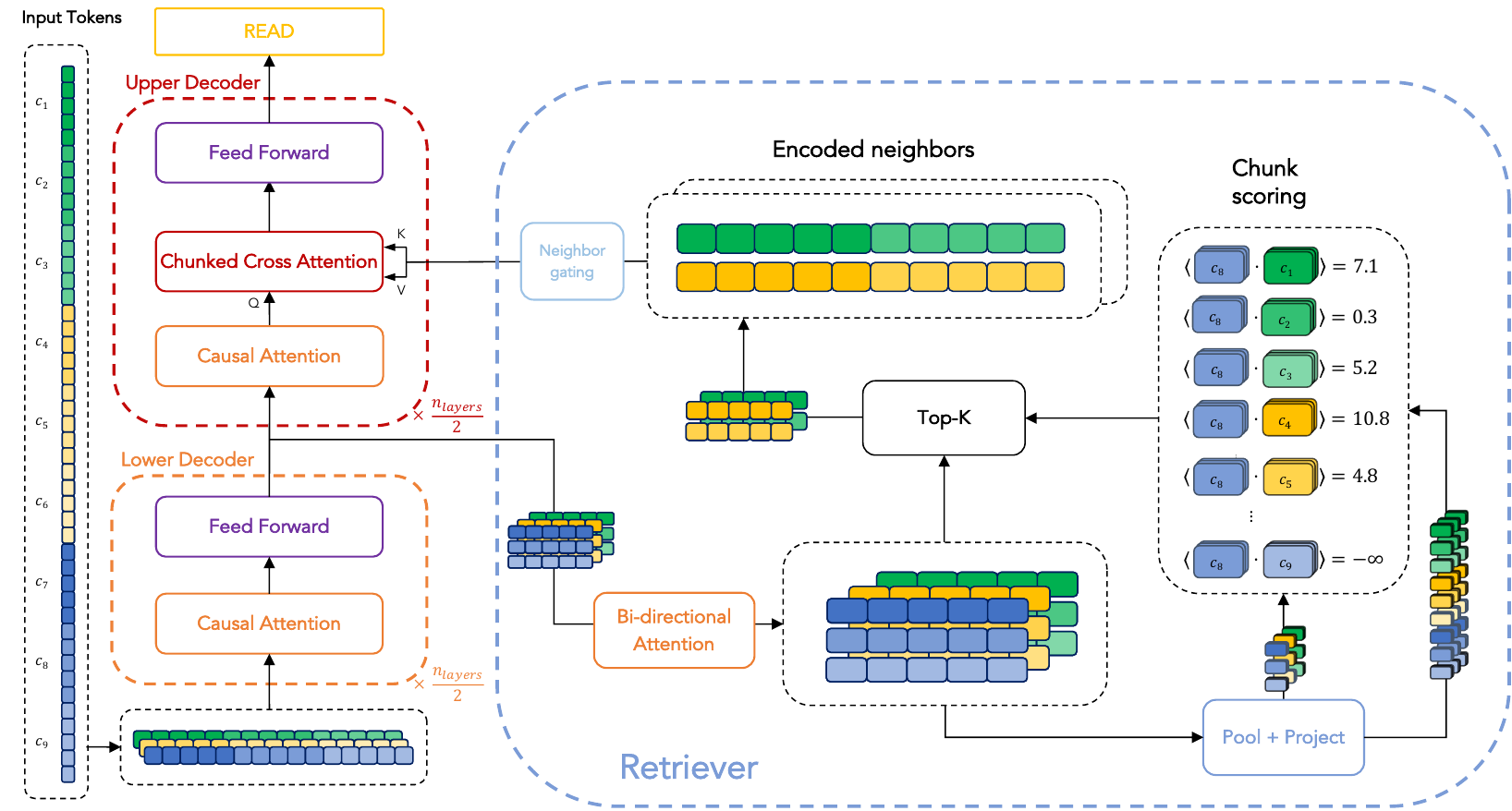}
    \caption{
 The architecture of the \textit{Retrieval-Pretrained Transformer}, where an input of 45 tokens is shown, consisting of 9 chunks, and causal self-attention is applied over 15 tokens. The left side shows the decoder stack, where the bottom $\frac{n_\text{layers}}{2}$ are standard Transformer decoder layers, and the top $\frac{n_\text{layers}}{2}$ layers also include chunked cross-attention layers that fuse information from retrieved chunks. The right side shows the retriever, which takes a chunk and retrieves the highest-scoring $K$ chunks that appeared earlier in the document.}
    \label{fig:high_level_architecture}
\end{figure*}

\section{Retrieval-Pretrained Transformer}
\label{sec:method}

\paragraph{Problem Setup}
Like RETRO \cite{retro}, RPT is a chunk-wise retrieval-augmented LM that divides the input sequence into chunks for retrieval. Specifically, given a sequence of $L$ input tokens, $\left(x_1, x_2, \dots, x_L\right)$, we partition it into a sequence of $\ell = \frac{L}{m}$ non-overlapping chunks of length $m$, denoted by $\mathcal{C} = \left(c_1, c_2, \dots, c_\ell\right)$. 
For every possible \emph{query} chunk, $c^{\textbf{q}} = c_i$, the 
model will retrieve a subset of at most $K \ll \ell$ chunks, $\mathcal{R}(c^{\textbf{q}}) \subset \mathcal{C}^{<i}=\left(c_1,c_2,...,c_{i-w}\right)$, where $\mathcal{C}^{<i}$ is the set of \emph{retrievable} chunks for $c_i$, which excludes the $w$ chunks to which it already has access to through causal self-attention. The goal is to learn a model that retrieves a chunk subset, $\mathcal{R}(c^{\textbf{q}})$, that increase the probability of autoregressive generation of the \textit{target chunk} $c^{\textbf{t}} = c_{i+1}$.

We present our method in two parts. First, our architecture (\S\ref{subsec:architecture}), which leverages CCA to fuse retrieved representations into the LM, but adds a learned retriever component. Second, we present the training method (\S\ref{subsec:supervision}-\S\ref{subsec:training}), where the retriever is trained to retrieve chunks useful for generating a future chunk according to a reference LM.

\subsection{Model Architecture}
\label{subsec:architecture}

Fig.~\ref{fig:high_level_architecture} illustrates our architecture, where the input has 45 input tokens divided into 9 chunks, and
causal self-attention is applied over $w=3$ chunks (15 tokens).
The left side depicts the decoder stack (\emph{``reader''}), and the right side the retriever. The reader is split into two, where the bottom $\frac{n_\text{layers}}{2}$ layers (\emph{lower decoder}) are standard Transformer decoder layers that take $w$ chunks as input and output representations that will be used by the retriever and the top decoder layers.

The top $\frac{n_\text{layers}}{2}$ layers (upper decoder)
use Chunked Cross-Attention (CCA) to fuse information from the top-$K$ neighbor chunks retrieved by the retriever back into the LM.
We use standard CCA layers from RETRO \cite{retro}, where 
for each one of the $\ell$ chunks, queries are the $m$ token representations of that chunk output by causal attention, and the keys and values are the token representations for the top-$K$ neighbor chunks output by the retriever.\footnote{For full details of CCA, see \newcite{retro}.}

Next, we describe the retriever component, along with a neighbor gating mechanism for modulating the effect of retrieved representations.

\paragraph{Retriever}
The retriever takes as input the representations output by the lower decoder and produces a similarity score for every pair of chunks. Given a \textit{query chunk} $c^\textbf{q}$, the \textit{query-based score}  for each retrievable chunk $c$ is $s_{\textbf{q}}( c)= \langle W_Q\textbf{c}^\textbf{q}, W_K\textbf{c} \rangle $, where $W_Q, W_K \in \mathbb{R}^{d \times d}$ are learned linear projections, and $\textbf{c}^\textbf{q}$ and $\textbf{c}$ are chunk representations.

For an $m$-token long chunk $c$, we compute its representation $\textbf{c}$ by applying bidirectional attention over the chunk tokens, followed by mean-pooling across the time dimension. This maintains causality, as these representations are only used during the prediction of the next chunk.

Once scores for all pairs of chunks are computed, the \emph{retrieved neighbor chunks} $\mathcal{R} (c^\textbf{q})$, for each query chunk, $c^\textbf{q}$, consists of its top-$K$ highest-scoring retrievable chunks. Then, for each chunk $c_j\in\mathcal{R}(c^\textbf{q})$, we concatenate the representations of the succeeding chunk $c_{j+1}$ to provide additional context, and the final representation for all neighbors of all chunks is given by a tensor $C \in \mathbb{R}^{\ell \times K \times 2m \times d}$.\footnote{Similar to RETRO, token representations of retrieved chunks are also augmented through cross-attention over tokens of the query chunk, $c^{\textbf{q}}$.}

Overall (and unlike methods like TRIME and kNN-LM), the retriever is an integral part of the LM, where the lower decoder computes representations for the retriever (which we dub \emph{self-retrieval}), and the upper decoder consumes representations produced by the retriever.

\paragraph{Neighbor gating} 
We add a neighbor gating mechanism to softly select neighbor representations that are useful for fusing into the upper decoder. 
Let $C_{i,k} \in \mathbb{R}^{2m \times d}$ be the token representations for the $k$'th neighbor of chunk $c_i$. We mean-pool across the time dimension to obtain a vector $\hat{\mathbf{c}}_{i,k}$ for each neighbor chunk. Then, we enrich the neighbor representation of each chunk by applying causal attention --  a neighbor chunk representations $\hat{\mathbf{c}}_{i,k}$ attends to chunks that precede it or to neighbors of the same chunk $c_i$ that are ranked higher. Finally, for each chunk we obtain the  \textit{gated retrieved representation} by multiplying the augmented representations by a gating score:
$C_{i,k}^{\textbf{g}} = \max\{\eta,\sigma(\frac{\mathbf{w}_\text{ng}\hat{\mathbf{c}}_{i,k}}{d})\} \cdot C_{i,k} $ where $\mathbf{w}_\text{ng}$ is a learned parameter vector,  $\eta$ is a small value meant to maintain gradient flow,\footnote{We set $\eta=0.1$ in all of our experiments.}  and $\sigma$ is the sigmoid activation.  Finally, in the upper decoder, when CCA is performed, the keys and values are $C_{i,k}^{\textbf{g}}$.

\subsection{Supervision Signal}
\label{subsec:supervision}

For each query chunk $c^{\textbf{q}}=c_i$, we want to identify neighbor chunks that will be helpful for generating $c^{\textbf{t}}=c_{i+1}$, and use those neighbor chunks as supervision signal for the retriever. Similar to \newcite{epr}, we can exploit the fact that we are producing \emph{training data} and use information from $c^{\textbf{t}}$ itself to produce such a score. Unlike \newcite{trime}, who use lexical clues alone, we will use an independent \emph{scoring LM} for this purpose.

Scoring every chunk w.r.t to all preceding chunks is quadratic in the number of chunks in a document, and thus computationally difficult. Thus, we use a simple, BM25 unsupervised retriever \cite{bm25}  \ that takes as input the concatenation of the chunks $(c^{\textbf{q}}, c^{\textbf{t}})=(c_{i}, c_{i+1})$ and returns a set of candidates neighbor chunks, $\bar{\mathcal{R}} \subset \mathcal{C}(c^\textbf{q})$, which have high lexical overlap with the current and subsequent chunk. This retriever has access to the tokens that need to be generated by the LM, which is allowed at training time.

Let $\hat{g}$ be an independently-trained LM, and let $\bar{c}_j$ be the concatenation $(c_{j}, c_{j+1})$. We compute a score $s_{\textbf{t}}\left(\bar{c}_j\right)$ 
that reflects whether the information in $\bar{c}_j$ is more useful for decoding $c^{\textbf{t}}$ compared to chunks that are close to $c^{\textbf{q}}$.
Specifically, the \textit{target-based score} for a candidate chunk is 
$$
s_{\textbf{t}}\left(\bar{c}_j\right)= \log\frac{\operatorname{Prob}_{\hat{g}}\left(c^{\textbf{t}} \mid c_{j}, c_{j+1}, c^{\textbf{q}}\right)}{\operatorname{Prob}_{\hat{g}}\left(c^{\textbf{t}} \mid c_{i-2}, c_{i-1}, c^{\textbf{q}}\right)}.
$$
This score is positive when information in $\bar{c}_j$ is more useful for decoding $c^{\textbf{t}}$ than information in the preceding two chunks $(c_{i-2}, c_{i-1})$.

We apply this scoring function to all chunks, and define for each query chunk $c^{\textbf{q}}$ the set of \emph{positive chunks} $\mathcal{R}_{\text{pos}}^{\textbf{q}}$, which includes candidates for which $s_{\textbf{t}}(\cdot) > 0$. 
This should result in helpful chunks, as each candidate chunk is at least as good as the local context. With this ordering at our disposal, we can apply standard retrieval training methods.

\subsection{Training}
\label{subsec:training}
To train the parameters of the retriever component, we adapt the widely-used LambdaRank loss \cite{lambdarank}. The loss for each query chunk $c^\textbf{q}$ (w.r.t its retrievable chunks) is:
\begin{align*}
&L_{\text{ret}}(c^{\textbf{q}})=\\
&\sum_{\{j,l: \bar{c}_l \in \mathcal{R}_{\text{pos}}^{\textbf{q}}, s_{\textbf{t}}(\bar{c}_l)>s_{\textbf{t}}(\bar{c}_j)\}  }\!\!\!\!\!\!\!\!\!\!\!\!\!\!\!\!\!\!\!\! \lambda_{jl}  \max \left(0,\tau-\left(s_{\textbf{q}}(c_l)-s_{\textbf{q}}(c_j)\right)\right)
\end{align*}
where $\tau$ is a margin hyper-parameter, and $\lambda_{jl}$ is the LambdaRank scaling that considers the relative ranking of each candidate. This loss is non-zero when for some pair of candidates, the target-based score disagrees (with margin $\tau$) with the ranking of the query-based score for candidates in $\mathcal{R}_{\text{pos}}^\textbf{q}$.
Optimizing this loss function allows RPT to distinguish between relevant and irrelevant chunks. 
Our final loss is $L_{\text{LM}}+\alpha_{\text{ret}}L_{\text{ret}}$, where $L_{\text{LM}}$ is the standard LM loss and $\alpha_{\text{ret}}$ is the retrieval loss coefficient, increased linearly in the first 100K steps. We also increase $\tau$ linearly during training.

\subsection{Important Implementation Details}
\label{sub:impo_imp}

\paragraph{Scheduled sampling}
To reduce train-test mismatch, we apply scheduled sampling \cite{scheduled_sampling} during training. Namely, after computing the top-$K$ neighbor chunks, we use these neighbors with probability $1-p_{\text{ss}}$, and with probability $p_{\text{ss}}$ the top-$K$ scoring candidates from $\mathcal{R}_{\text{pos}}^\textbf{q}$ as input for CCA. We anneal $p_{\text{ss}}$ from 1 to 0 during the first 90\% of training with a cosine schedule. This allows the model to gradually learn to use its own predictions. We report the effect of this in \S\ref{sub:ablations}.

\paragraph{Sliding window attention at training and inference time} 

As described in \S\ref{sec:method}, the decoder takes as input $w$ chunks, each with $m$ tokens as input, and applies causal attention over them. In practice, to give the first tokens access to past tokens, we use the sliding-window attention mechanism \cite{txl,longformer,SLED}, where the number of tokens in a window is 2,048 and the stride is 1,024. Thus, the input to each window is 2,048 tokens and the output are the representations for the last 1,024 tokens, which use the keys and values of the previous 1,024 tokens for contextualization.

At inference time a similar procedure is applied. We compute and cache the key and value representations for segments of 1,024 tokens, using these as context for generating or estimating the probability of the next segment.

\paragraph{Retrieval at inference time} \new{During training we encode in each batch sequences of length 16K and retrieve chunks from those encoded 16k tokens. However, at inference time the retriever provides access to \emph{all} tokens from the start of the document, where we store the key and lower-decoder representations in a Faiss~\cite{faiss} index on the CPU. For each chunk, we query the index using the chunk's query representations and retrieve the top-$K$ lower-decoder representations with the highest dot product.}

\paragraph{Additional details}
At training time we use sequences of length $L=16,384$ tokens, which are split into 4 devices, each consuming $4,096$ tokens. As mentioned, the decoder stack takes $2,048$ tokens as input (in a sliding window approach), which contains $\ell=32$ chunks of length $m=64$. We employ Rotary Positional embedding \cite{rotary}, and train all models for 500K steps on a TPUv4-64, with an effective batch size of ${2^{17}}$ tokens resulting in a total training budget of 65 billion tokens.  

For all models trained, we use the GPT-NeoX \cite{neox} tokenizer, which was trained on the Pile \cite{pile} and covers the domains we evaluate on (see \S\ref{sec:datasets}).
As our scoring language model, we use the deduplicated 1.4B parameter version of Pythia \cite{pythia}, and score with it the top-20 BM25 candidates.
Our model has 12 layers, hidden dimension $d=1024$, and 8 attention heads with a head dimension of 128. We apply CCA with 2 neighbors, unless mentioned otherwise.
Additional implementation details are in Appendix~\ref{app:additional_imp} and theoretical complexity of CCA layers is in Appendix \ref{app:complexity}.

\begin{figure}[h]
    \centering
    \includegraphics[width=0.48\textwidth]{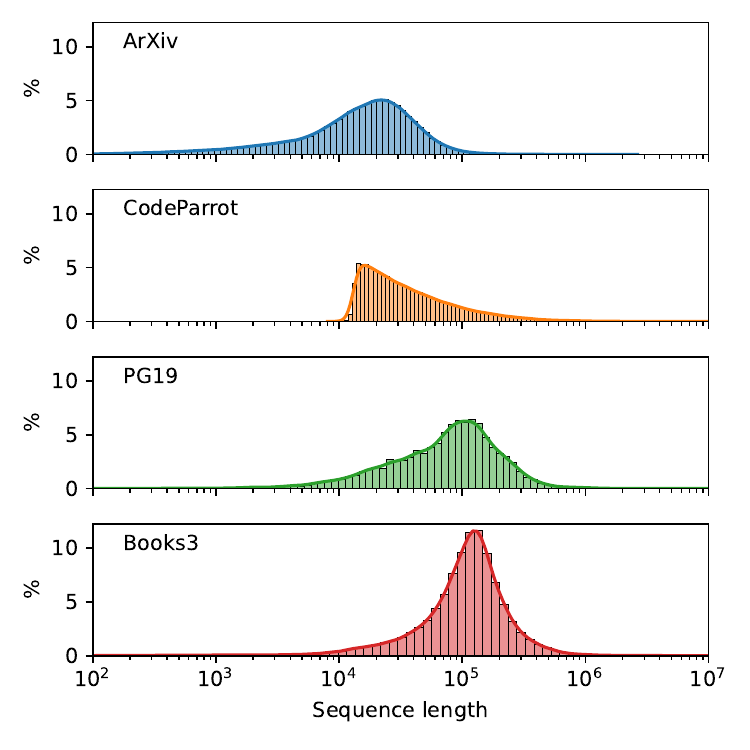}
    \caption{Histograms of the distribution over document length in tokens across all datasets. The x-axis is in log scale.
    }
    \label{fig:tok_hist}
\end{figure}

\begin{table}[]
\centering
\scalebox{0.9}{
{\footnotesize
\begin{tabular}{ccc}
\toprule
\textbf{Name} & \textbf{Tokens (Train/Test)} & \textbf{Median Length} \\
\midrule
\textbf{ArXiv} & 12,000 / 16  & 16,368 \\
\textbf{CodeParrot} & 5,000 / 5 & 29,269 \\
\textbf{PG19} & 3,000 / 9 & 82,659  \\
\textbf{Books3} & 25,000 / 35 & 113,496\\

\end{tabular}
}}
\caption{Number of tokens (in millions) for each dataset and median document length.}
\label{table:dataset_token_split}
\end{table}


\section{Long Range LM Datasets}
\label{sec:datasets}

We evaluate RPT on four datasets, covering domains such as books, code, and mathematical writing, which require the ability to recall information over long distances. 
Tab.~\ref{table:dataset_token_split} and Fig.~\ref{fig:tok_hist} provide statistics on dataset size and the distribution over document length, showing that documents are long across all datasets and in particular PG19 and Books3, where documents typically contain $10^5$ tokens or more. We briefly review the datasets.

\paragraph{PG19}
\label{subsec:pg19}
Introduced in \newcite{Rae2020Compressive}, PG19 is a widely-used long-range language modeling benchmark containing books from Project Gutenberg, and covering a wide range of literary genres, styles, and topics. 
We adopt the exact setup and data split from prior work \cite{memorizing,blockrecurrent,gss}.

\paragraph{Books3}
\label{subsec:books3}
is a corpus of books released as part of the Pile \cite{pile}, containing a vast collection of literary works from different domains. To our knowledge, we are the first to use this corpus as a long-range language modeling benchmark.\footnote{We do not release this benchmark due to the copyright restrictions.}

\paragraph{CodeParrot} \cite{codeparrot} is a corpus of clean, nearly-deduplicated Python code from various GitHub repositories. Modeling code requires understanding patterns and contextualizing information over long distances, making it a natural candidate for testing long-range LMs. In our experiments, we follow the approach of \newcite{memorizing}, combining files from the same repository to construct a corpus with longer sequences, and create a train/test split (see Tab.~\ref{table:dataset_token_split}).

\paragraph{ArXiv}
\label{subsec:arxivmath}
is a corpus of preprint papers extracted from ArXiv. It consists of mathematical texts that require maintaining coherence and referring to previously mentioned information over extended text. 
Prior work evaluated long-range LMs on this corpus \citep{memorizing,blockrecurrent,gss}, but did not release their corpus. Thus, we use the preprocessed corpus and data splits made available by \newcite{proofpile}.

\section{Experiments}

We now turn to experiments for comparing RPT to prior work across our four datasets.

\begin{table*}[t]
\centering
\scalebox{0.95}{
{\footnotesize
\begin{tabular}{lllllll}
\toprule
\textbf{Model}                        & \textbf{ArXiv} & \textbf{Code} & \textbf{PG19} & \textbf{Books3} & \textbf{Params} & \textbf{Time/update} \\
\toprule
\textsc{Transformer-xl (our impl.)}     & 3.11       & 2.30            & 11.48     & 15.00    & 202M & 1$\times$           \\
\quad\textsc{+2 layers}     & 3.07       & 2.26            & 11.2     & 14.52    & 228M & 1.14$\times$           \\
\quad\textsc{1.5$\times$ additional steps}     & 3.11 & 2.26 &  11.39 &    14.70    & 202M & 1$\times$           \\
\textsc{Retro w. BM25 (our impl.)}              & 2.94       & 2.17           & 11.44     & 14.60 &  236M & 1.35$\times$              \\
\textsc{RPT-Lex} & 2.92 & 2.23                   & 11.59             & 14.32      & 242M & 1.51$\times$         \\ 
\textsc{RPT-Sem}  & 2.77       & 2.17          & 10.96   & 13.91 & 242M & 1.51$\times$     \\   
\quad\textsc{w. 3 neighbours}  & 2.75       & 2.16          & \textbf{10.92}   & \textbf{13.87}   &  &    \\   
\quad\textsc{w. 4 neighbours}  & \textbf{2.74}       & \textbf{2.15}          & 10.93   & 13.91  &   &    \\   
\textsc{Memorizing Transformer (32K) }     & 2.92       & 2.18            & 10.97     & 14.40    & 212M   & 1.82$\times$        \\
\textsc{Memorizing Transformer (65K) }     & 2.93       & \textbf{2.15}            & 10.99     & 14.3    & 212M   & 2.12$\times$        \\
\textsc{Block-Recurrent Transformer}    & {2.89}       & 2.73            & {10.95}     & 14.64    & 212M  & 1.56$\times$        \\
\textsc{Griffin}     & 3.08 & 2.24 &  11.26 &    14.16    & 240M & 1.15$\times$           \\


\midrule 
\textsc{RPT-Lex w. Oracle}  & 2.80       & 2.12          & 10.88   & 13.30     & 242M  & 1.51$\times$    \\
\textsc{RPT-Sem w. Oracle}  & 2.69       & 2.10          & 10.26   & 12.74      & 242M   & 1.51$\times$ 

\end{tabular}
}}
\caption{Test set perplexity for all datasets along with number of parameters and the relative increase in time per update during training compared with Transformer-XL. Unless specified, models are trained for 500k steps and use 2 neighbours during inference.}
\label{tab:results}
\end{table*}

\subsection{Experimental Setup}
We compare to the following baselines and oracles.

\paragraph{Transformer-XL} Our simplest baseline is a standard transformer decoder stack with sliding window attention. Put differently, we simply remove from RPT the retriever component and CCA layers in the upper decoder. Using sliding window attention (as described in \S\ref{sub:impo_imp}) can be viewed as a variant of Transformer-XL \cite{txl}.
\new{We compare RPT to Transformer-XL in multiple settings, one where we have the same number of layers and training steps for both models, and two more where we tie the number of parameters and FLOPs between the models.}


\paragraph{RETRO} \new{We implement a modified version of \newcite{retro}}, a retrieval-augmented model, where feed the top-$K$ neighbors retrieved by BM25\footnote{Concurrent work \cite{surfacebased} showed that training RETRO using BM25 outperforms dense retrieval methods.} as input to the CCA layers in the upper decoder.  \new{Concretely, \newcite{retro} performed CCA over the representation from a separate bi-directional encoder, while our variant uses the lower-decoder representations as a replacement. This makes RPT and RETRO architectures more similar to one another and allows evaluation to center on the importance of training the retriever, which is the focus of our work.}
During training, we use the query $(c^{\textbf{q}},c^{\textbf{t}})$, since we have access to the target chunk. During inference, we use $c^{\textbf{q}}$.

\paragraph{RPT-Lex} 
A version of RPT, where the training signal is obtained solely from lexical information, similar to TRIME \cite{trime}. Explicitly, the set of positive chunks $\mathcal{R}_{\text{pos}}^\textbf{q}$ for a chunk $c^{\textbf{q}}$ contains the top-20 chunks that have the highest BM25 score with $(c^{\textbf{q}}, c^{\textbf{t}})$.

\paragraph{RPT-Sem} Our full model described in \S\ref{sec:method}.

\paragraph{Block-Recurrent Transformer} 
We use the official training implementation\footnote{\label{note1}\url{https://github.com/google-research/meliad}.} of Block-Recurrent Transformer \cite{blockrecurrent} with the default configuration. 

\paragraph{Memorizing Transformer}
We use the official implementation\footref{note1} of Memorizing Transformers \cite{memorizing}, with the default configuration and a memory size of 32K and 65K tokens. 

\paragraph{Griffin}
An alternative for long-range modeling is to use a hybrid of attention and linear RNNs \cite{lru, dlr}. We evaluate Griffin \cite{griffin}, a state-of-the-art model in this category. We adapt the official implementation, and supplement our Transformer-XL baseline with 5 recurrent layers in the final layers to ensure parameter parity. We use a state dimension of 2,048, and temporal dimension of 3.

\paragraph{Oracles}
For each test chunk, we can exhaustively search and use at test time the best possible neighbors for a model according to the scoring LM.
This provides an upper bound for the performance of RPT-Sem, as it is trained to imitate the ranking produced by this oracle.

\paragraph{Metrics} 
We use perplexity to evaluate the performance of models. 
In addition, we use the target score $s_\textbf{t}(\cdot)$ from the scoring LM to compute for each chunk a gold ranking over all previous chunks, and to label chunks as positive/negative iff their target score is positive/negative, respectively.
With this information, we can evaluate Precision@$k$, which is the fraction of top-$k$ chunks according to the query-based score that are positive, and Recall@$k$, which is the fraction of positive chunks that are in the top-$k$ chunks according to the query-based score. We also use the gold ranking to compute NDCG@$k$, which is a standard retrieval metric \cite{ndcg}.

\subsection{Results} 
Table \ref{tab:results} shows our main results, which show that RPT-Sem is comparable or better than all other baselines in all cases. Using a fixed retriever (RETRO) improves performance compared to Transformer-XL; RPT-Lex leads to gains in Books3 but to losses in PG19 compared to RETRO, and RPT-Sem outperforms Transformer-XL, RETRO, and RPT-Lex on ArXiv, PG19, and Books3, and has performance comparable to RETRO on CodeParrot.
\new{Even in the parameters-tied and compute-tied setting, Transformer-XL still performs substantially worse than RPT.}
Compared to Block-Recurrent Transformer, Memorizing Transformers and Griffin, which do not use CCA, performance is again similar or better, with significant improvements on ArXiv and Books3.

CCA enables to dynamically increase the number of neighbors at inference time. When using 3 or 4 neighbors (instead of 2), performance improves, which allows compute-performance trade-offs.


Last, oracle models consistently achieve the best perplexity across all datasets, improving from 2.74$\rightarrow$2.69 on ArXiv, 2.15$\rightarrow$2.10 on CodeParrot, 10.92$\rightarrow$10.26 on PG19, and 13.87$\rightarrow$12.74 for Books3. This shows that improving retriever training can further improve performance.

\begin{table}[ht]
\centering
\scalebox{1.0}{
{
\tiny
\setlength{\tabcolsep}{1.8pt}
\begin{tabular}{lccc|ccc|ccc}
\toprule

   \textbf{Dataset} & \multicolumn{3}{c|}{\textbf{Precision@2}} & \multicolumn{3}{|c|}{\textbf{Recall@10}} & \multicolumn{3}{|c}{\textbf{nDCG@20}} \\
           & \textsc{BM25} & \textsc{RPT-L} & \textsc{RPT-S} & \textsc{BM25} & \textsc{RPT-L} & \textsc{RPT-S} & \textsc{BM25} & \textsc{RPT-L} & \textsc{RPT-S} \\
\midrule
     \textbf{ArXiv} &       27\% &    26\% &    32\% &       55\% &    54\% &    58\% &       24\% &    24\% &    30\% \\
\textbf{Code} &       29\% &    26\% &    34\% &       53\% &    52\% &    56\% &       25\% &    23\% &    30\% \\
      \textbf{PG19} &       22\% &    22\% &    28\% &       55\% &    55\% &    61\% &       18\% &    18\% &    23\%  \\
    \textbf{Books3} &       23\% &    19\% &    26\% &       55\% &    50\% &    58\% &       18\% &    16\% &    22\% \\
    \midrule
    \textbf{Avg} & 25.2\% & 23.2\% & \textbf{30.0\%} & 54.5\% & 52.7\% & \textbf{58.2\% }& 21.2\% & 20.2\% & \textbf{26.2\% }
\end{tabular}
}}
\caption{Test retrieval metrics across datasets.
}
\label{tab:ret_metrics}
\end{table}

\paragraph{Retrieval metrics}
 Table \ref{tab:ret_metrics} presents the retrieval metrics w.r.t oracle positive chunks. 
 Again, retrieval with RPT-Sem outperforms both RPT-Lex and BM25 in all cases. This shows the importance of training a retriever, and moreover that using semantic supervision leads to better retrieval compared to a lexical signal only.



\subsection{Ablations}
\label{sub:ablations}

\begin{table}
\centering
\scalebox{0.75}{
{\footnotesize
\begin{tabular}{lllll}
\toprule
\textbf{Model}                         & \textbf{ArXiv} & \textbf{Code} & \textbf{PG19} & \textbf{Books3} \\
\toprule

\textsc{Retro w. BM25 (our impl.)}              & 2.94       & 2.17           & 11.44     & 14.60              \\
\quad\textsc{w. DPR-style retriever}  &  2.97      &     2.28     &  11.7  &  14.86    \\   
\textsc{RPT-Lex} & 2.92 & 2.23                   & 11.59             & 14.32               \\ 
\quad\textsc{w. DPR-style retriever}  & 2.84       & 2.26         & 11.11   & 14.17    \\   
\textsc{RPT-Sem}  & 2.77       & 2.17          & 10.96   & 13.91      \\   
\quad\textsc{w. DPR-style retriever}  & 2.98       & 2.33         & 11.62   & 14.66  \\
\textsc{RPT-Sem - Only Teacher forcing} & 2.91              & 2.22                   & 11.54             & 14.66           \\
\textsc{RPT-Sem - No Teacher forcing}   & 2.95              & 2.26                   & 13.10             & 14.40          \\
\textsc{RPT-Sem - No Neighbor Gating}            & 2.92              & 2.20                   & 11.50             & 18.68    \\   

\end{tabular}
}}
\caption{Results of our ablation study. }
\label{tab:ablations}
\end{table}




Tab.~\ref{tab:ablations} shows the result of an ablation study over all datasets.

\paragraph{Only Teacher Forcing} 
We force the model to attend to gold neighbors according to the scoring LM, without annealing $p_{\text{ss}}$ during training. This leads to a performance drop across all datasets, and in particular for PG19 and Books3.

\paragraph{No Teacher Forcing} 
Here, we do the opposite and fix $p_{\text{ss}}=0$ throughout training, i.e., we only use the predicted neighbors and not gold ones. This can lead to undertraining of the CCA layers since they are exposed to low-quality neighbors at the beginning of training and results drop even further compared to Only Teacher Forcing.

\paragraph{No neighbor gating} We disable neighbor gating which controls the flow of information from neighbor chunks and analyze the effect on model performance. We observe a performance reduction across all datasets, notably on Books3, where perplexity increases by 4.5 points.

\paragraph{DPR-style retriever}
\new{
To study the importance of joint training, we test performance when using retrievers that are trained separately from the LM, thereby inducing a train-test mismatch.
We train dense retrievers using the standard DPR training procedure \cite{dpr} on each dataset (see Appendix~\ref{app:dpr_details} for training details), and for each of our CCA models use this retriever instead of the one it was trained with. 
Interestingly, we observe RPT-Lex can effectively utilize the DPR-style neighbors giving it a slight performance improvement on 3 of the 4 datasets. 
}

\new{
As expected, the two models trained with the stronger retrievers suffer from the train-test mismatch, replacing the BM25 retriever and RPT-Sem retriever with the DPR-style retriever causes both models to suffer performance degradation on all datasets, suggesting that the non-ablated performance is the result of coordination between the retriever and the language model.
}

\subsection{Analysis}
\label{sub:analysis}
\begin{figure}[h]
    \centering
    \includegraphics[width=0.48\textwidth]{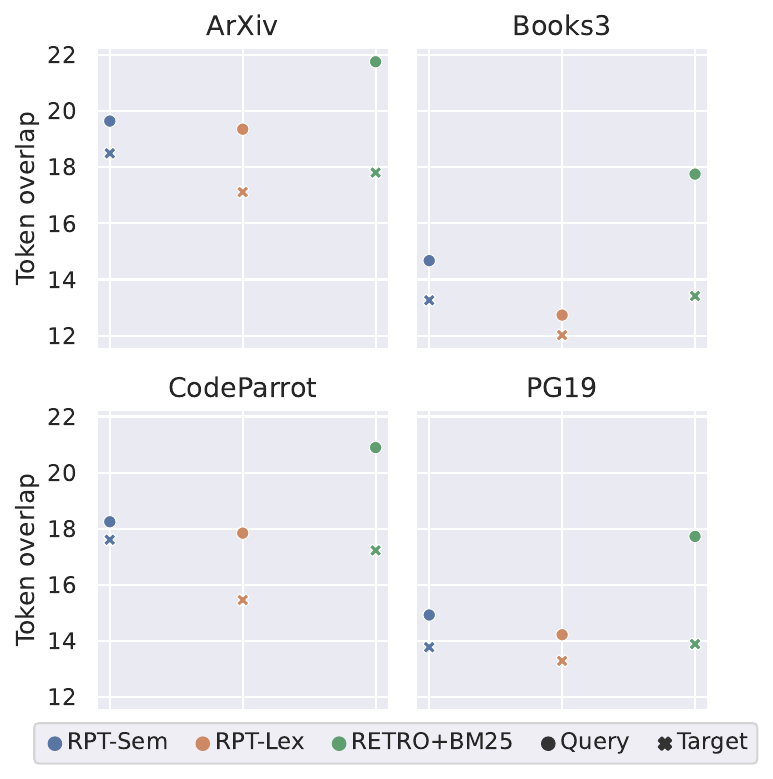}
    \caption{We measure the number of unique token overlap between query/target chunks and the best retrieved neighbor.
    }
    \label{fig:overlap_w_curr}
\end{figure}

\paragraph{Token overlap} 
Fig.~\ref{fig:overlap_w_curr} plots the average number of tokens that overlap between the query/target chunks in the best retrieved neighbor for RETRO, RPT-Lex, and RPT-Sem. RPT-Sem retrieves paragraphs with higher overlap with the \emph{target} chunk compared to RPT-Lex. Naturally, BM25 retrieves chunks with the highest overlap with the \emph{query} chunk. However, this does not translate to higher lexical overlap for the \emph{target} chunk.

\begin{figure}[h]
    \centering
    \includegraphics[width=0.47\textwidth]{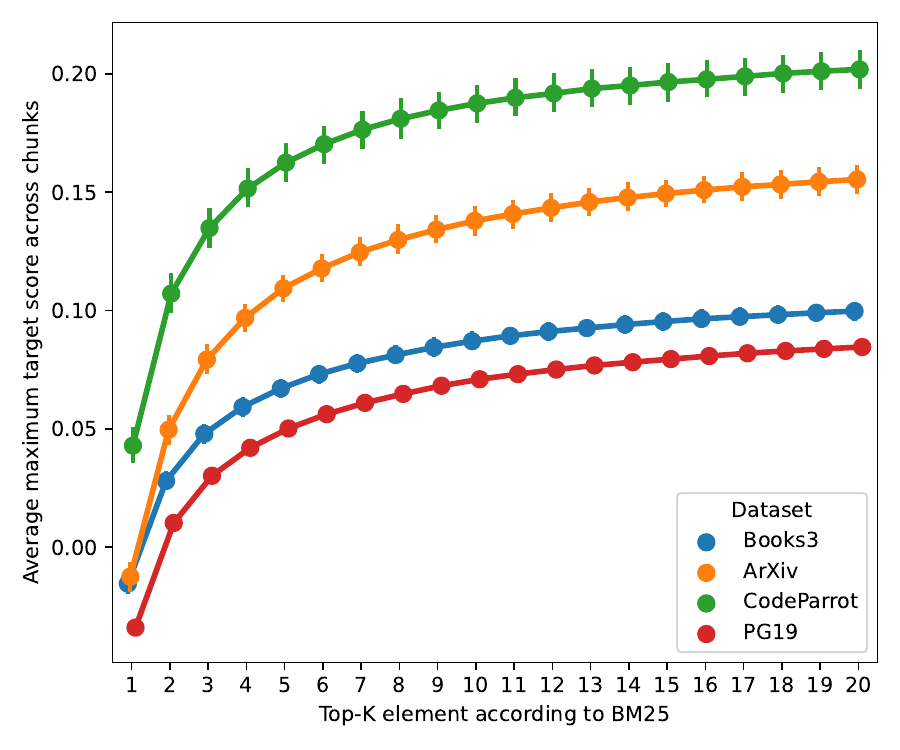}
    \caption{
    The maximal target score $s_\textbf{t}(\cdot)$ for the top-$K$ chunks retrieved by BM25 averaged across chunks and for all datasets. Since the maximal target score for the top-20 chunks is much higher than for the top-2, 
    learning to rerank the top-20 BM25 candidates can lead to substantial improvements in retrieval quality.
    }
    \label{fig:scored_ds}
\end{figure}

\paragraph{Supervision quality}
We train RPT-Sem using information from the target scoring function $s_\textbf{t}(\cdot)$, which we saw leads to model improvements.
However, the target scoring function only provides a reranking of the top-20 candidates according to BM25. Thus, a natural question is how much does the supervision quality improve through this reranking. Fig~\ref{fig:scored_ds} shows for every rank $K$ the maximal target score among the top-$K$ chunks according to BM25, averaged over chunks and across our 4 datasets. Clearly, reranking the top-20 BM25 candidates has a lot of potential, as the maximal target score is much higher for the top-20 candidates compared to the top-2. This hints that longer and better training of the retriever can further improve the performance of RPT-Sem.

Interestingly, our analysis sheds light on why RPT-Sem outperforms RETRO clearly on Books3 and PG19 but less so on CodeParrot. The maximal target score for CodeParrot when $k=2$ is already quite high -- around 0.1, which corresponds to more than 10\% improvement in the probability of the target chunk compared to the local context. Conversely, for PG19 and Books3, the target score when $k=2$ is closer to 0.

\begin{figure}[h]
    \centering
    \includegraphics[width=0.48\textwidth]{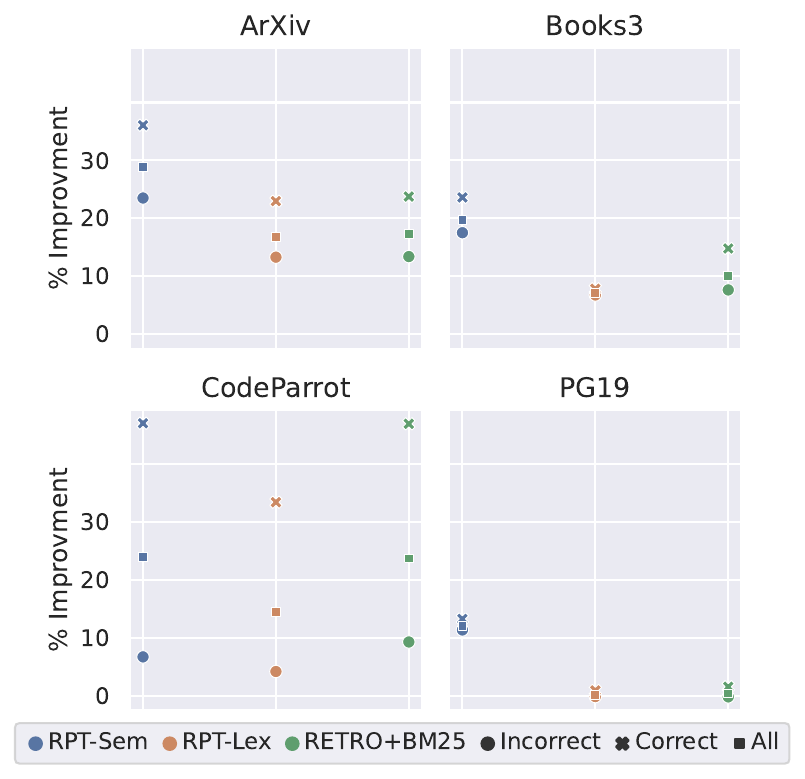}
    \caption{Relative improvement with/without correct retrieval.
    }
    \label{fig:imp_w_wo_corr_pred}
\end{figure}

\paragraph{Subgroup analysis} 
Fig~\ref{fig:imp_w_wo_corr_pred} shows the average relative improvement (across chunks) of RETRO, RPT-Lex, and RPT-Sem compared to Transformer-XL, when distinguishing between cases where a ``gold'' oracle chunk was retrieved and cases where no gold chunk was retrieved. 

As expected, RPT-Sem leads to improvements on all datasets, and outperforms other baselines except for RETRO on CodeParrot where performance is similar.
Second, cases where a gold chunk was retrieved indeed typically lead to larger improvements, but we witness improvements even in cases where a gold chunk was not retrieved, which shows that the model can still benefit from such retrievals.

\paragraph{Qualitative analysis} \new{Examining retrieved chunks, we observe that the RPT retriever is highly contextual. When applied on code, it retrieves function definitions, variable assignments, etc., on ArXiv it retrieves definitions of lemmas, theorems, etc.} Fig.~\ref{fig:quan1} shows an example, where we give the codebase used for this paper as input to our model and present an example query chunk where RPT produces better retrieval than BM25. We observe that the preceding context allows RPT to effectively retrieve a relevant object definition, leading to lower loss. 



\begin{figure*}[h]
    \centering
    \begin{adjustbox}{scale=0.8}
    \input{examples/quant_2}
    \end{adjustbox}
    \caption{
    An illustrative example showcasing the top-1 retrieved neighbors for both \PY{sem}{\textbf{RPT-Sem}} and \PY{bm25}{\textbf{BM25}} models applied to RPT's code. The variable \texttt{outputs} in the \PY{query}{\bf{query chunk}} is a member of the class \texttt{FlaxRPTRetrieverEncodedOutput}. \PY{sem}{\textbf{RPT-Sem}} successfully retrieves the object's definition leading to a reduced loss on the \PY{target}{\textbf{target chunk}}, in comparison to \PY{bm25}{\textbf{BM25}}. 
    }
\label{fig:quan1}
\end{figure*}

\section{Discussion and Related Work }
\label{sec:related_work}

\paragraph{Relation to Fusion-in-Decoder} \new{RPT shares similarities with Fusion-in-Decoder (FiD) \cite{fid,SLED}. While both RPT and FiD employ cross-attention mechanisms to integrate the retrieved context within their models, they differ in two ways.
(a) In FiD, retrieval is performed only once based on the initial prompt/query, while RPT continuously performs retrieval at the chunk level throughout generation. (b) FiD encodes retrieved neighbors separately using a bi-directional encoder and only then applies cross-attention in the decoder. In RPT, the decoder computes chunk embeddings and performs native retrieval, and then chunked cross-attention is applied to fuse the retrieved context with the model's predictions. 
We view RPT, which uses lower-decoder encodings, as more natural in the context of continuous generation (e.g., chatbots or agents), since the model generates representations and uses them later as keys, and thus generating retrieval representations bears zero cost.}

\paragraph{Long-range language modeling}
A primary focus in long-range language modeling has been addressing the quadratic complexity of attention in order to develop more efficient mechanisms for handling long texts. For instance, Transformer-XL \cite{txl} processes the input using a segment-level mechanism while retaining a cache from previous segments. Longformer \cite{longformer} extends this idea to accommodate even longer contexts. \new{Several works previously viewed retrieval as a long-range problem. 
Memorizing Transformers \cite{memorizing} employed a single $k$-NN layer and retrieve cached keys and values, but they do not back-propagate gradients through the sparse retrieval operation. Similarly, \newcite{unlimiformer} demonstrated that this approach can be used with any existing pre-trained model and applied it at every attention layer for long summarization tasks.}
  From an analysis perspective, past work \cite{shortformer} demonstrated that standard LM benchmarks are not ideal for measuring the long-range capabilities of models. \newcite{DoLL} discuss various types of sequences that benefit from having a long context, and \newcite{do-transformers} investigate long-range architectural choices and recommend increasing long-range capabilities in the upper layers.

\paragraph{Efficient language modeling}  \new{Sparse strategies, such as those proposed in \newcite{bigbird,routing_transformers,reformer}, similarly to RPT, attend to only a subset of tokens through clustering or hashing methods, which are trained by propagating gradients through the sparse operation. In RPT, sparsity is due to the retriever top-K operation, which is trained using high-quality supervision from a reference language model. Another approach for efficiently modeling long text involves compressing the input and attending over the compressed sequence \cite{infinityformer,Rae2020Compressive}, or learning to ignore irrelevant tokens \cite{expirespan}. 
However, empirically most efficient transformer architectures trade off efficiency for quality. Recently, state-space models \cite{gss,mamba,h3} models emerged as an efficient alternative, which approaches Transformer quality. In this paper, we explore models that are based on classic quadratic Transformer. We argue 
that the underlying model is orthogonal to our contribution and can be replaced by other efficient alternatives and combined with retrieval. We leave this exploration for future work.}






\paragraph{Retrieval augmented LMs}
Retrieval-augmented LMs have emerged as a prominent approach for efficiently leveraging external knowledge while generating text. These models can be broadly divided into those operating at token-level granularity and those operating at sequence-level granularity. Token-level methods, such as kNN-LM \cite{knnlm}, TRIME \cite{trime}, and SPALM \cite{spalm}, retrieve information for individual tokens. Sequence-level approaches like RAG \cite{rag} utilize pre-trained encoder-decoder models with pre-trained retrievers for tasks like open-domain question answering. Similarly, FiD \cite{fid} employs generative encoder-decoder models that fuse evidence from multiple passages during the decoding process, closely related to the CCA mechanism.
Recently, \newcite{shall} demonstrated the potential benefits of conducting retrieval and chunked cross-attention at each time step, compared with the original RETRO \cite{retro} paper, which retrieves every $m=64$ steps.

\paragraph{Joint retriever-reader training}
 Joint training approaches typically concentrate on transferring information between a pre-trained reader into a pre-trained retriever. These methods commonly involve updating the retriever index during the training process in the context of knowledge-intensive tasks, such as open-domain question answering. For instance, REALM \cite{realm} utilizes masked language modeling as a learning signal to update the retriever. EMDR2 \cite{emdr2} extends FiD by using encoder-decoder models to back-propagate errors from the predicted answer to the retriever. Similarly, \newcite{fid2, reatt} uses attention scores from the reader to supervise the retriever directly using the attention matrix as a training signal to enable joint end-to-end training with the supervision of the downstream task.
Notably, \newcite{atlas} further scale up these approaches and jointly train a retriever with an encoder-decoder model, demonstrating strong few-shot learning capabilities. They also investigate various retriever updating techniques to address train-test mismatches in the retrieval process.
We do not encounter the issue of index update since we compute the entire index through a forward pass.



\paragraph{Retriever Pre-training}
Early work on retriever pre-training relied on the unsupervised Inverse Cloze Task to pre-train the retriever \cite{latent-lee,realm}. It was later shown that directly using BERT \cite{bert} with a supervised objective is sufficient to get good performance on standard benchmarks \cite{dpr}. However, this paradigm showed lackluster performance on long-tail entities compared to BM25 \citep{qampari,entityquestions}.
Recently, unsupervised pre-training methods \cite{cocondenser,spider,contriever} enabled improved performance. However, these methods are initialized from a pre-trained BERT \cite{bert} encoder model, while RPT is a retriever-reader architecture trained from scratch that outperforms BM25 without any additional pre-training.

\paragraph{Supervising retrievers with LLMs}
EPR \cite{epr} demonstrated that LLMs could be employed to train a retriever for prompt retrieval by estimating the probability of an output given the input and a candidate training example as the prompt. Similar techniques were applied to open-domain question answering via re-ranking retrieval results \cite{UPR,ram2023incontext} and to supervise retrievers through perplexity distillation \cite{atlas}. Recently, \newcite{replug} utilized this supervision method to improve the performance of various LLMs in a black-box fashion.

\section{Conclusion}
In this work, we present the Retrieval-Pretrained Transformer (RPT), a retrieval-augmented LM where the retriever is trained as a native component of the LM to retrieve semantically relevant chunks for future text prediction.  We evaluate RPT on four long-range language modeling tasks, including books, code, and mathematical writing. We demonstrate that by seamlessly integrating the retriever into the architecture and training process, RPT benefits from the fusion of retrieved context, improving over strong retrieval-augmented baselines. \new{While this work focuses on retrieval from long texts, we argue our empirical findings show that 
adapting our procedure for general web-based corpora retrieval is an exciting future direction. This will require overcoming technical difficulties related to scaling and pretraining corpus construction.} We envision RPT will pave the way for a new generation of pretrained language models with retrieval deeply integrated throughout their architecture and training process.







\section*{Acknowledgments}
This research was supported with Cloud TPUs from Google's TPU Research Cloud (TRC) and The European Research Council (ERC) under the European Union Horizons 2020 research and innovation programme (grant ERC DELPHI 802800). Ohad would like to thank Iz Beltagy for suggesting the TRC program, and the entire TAU NLP lab and especially Guy Dar and Itay Itzhak. This work was completed in partial fulfillment of the Ph.D. degree of Ohad Rubin.

\bibliography{anthology,custom, journals}
\bibliographystyle{acl_natbib}

\appendix
\section{Additional Implementation Details}
\label{sec:appendix}
\label{app:additional_imp}

Models are implemented in JAX with a dropout rate of 0.05, and the AdaBelief \cite{adabelief} optimizer with a weight decay of 1e-8, cosine decay to 0.1 of max learning rate, global gradient norm clipping of 1, and tied input embedding \cite{weight_tying}.  Grid search determined $\tau$ values: 128 for Books3, 4 for PG19, 2 for CodeParrot, and 8 for ArXiv. We set $\alpha_{\text{ret}}=1e-9$ for all datasets and a base learning rate of $5e-3$, using the validation set for hyperparameter selection.


\section{Computational Complexity}
\label{app:complexity}
\new{
The per token computational complexity of an attention layer in a transformer model with dimension $d$, $|Q|$ queries and $|K|$ keys is \( 2 \cdot d \cdot (|K| \cdot |Q| + |K| \cdot d + |Q| \cdot d) \) flops.\footnote{For a query matrix \( Q \in \mathbb{R}^{|Q| \times d} \) and a key/value matrix \( K \in \mathbb{R}^{|K| \times d} \), it consists of the following operations: multiplication with \( W_Q \), \( W_K \), and \( W_V \) for the queries, keys, and values, each costing \( |Q| \cdot d^2 \), \( |K| \cdot d^2 \), and \( |K| \cdot d^2 \) flops respectively. Computing the attention matrix and multiplying it by the values each requires \( |Q| \cdot |K| \cdot d \) flops. Finally, multiplying by the output matrix is an additional \( |Q| \cdot d^2 \) flops.} By setting $N=|Q|=|K|$ and adding the cost the feed-forward layer, we get that the per token cost for a transformer block when $d\gg N$ is $ 2d (N + 2d) + 8d^2 \approx 12d^2$ flops. 
For CCA, the cost is dependent on the chunk size $C$, and number of neighbors $k$. Setting $|K|=2Ck$ and $|Q|=C$, and assuming $d \gg Ck$, the cost per token for a CCA layer is $2d(2 C k + 2 d k + d)\approx (4k+2) \cdot d^2$ flops. Our per token overhead for $\alpha\in[0,1]$ of the blocks including CCA is $\approx \alpha (\frac{k}{3}+\frac{1}{6})$. In our experiments, we use CCA in 5 of the 12 layers so $\alpha=\frac{5}{12}$ and $k=2$, and get that CCA contributes an overhead of approximately 1.29x. Using similar logic, the constant cost for the retriever component is the two linear projections, the two additional bidirectional attention layers, and the query augmentation layer resulting in $\frac{1}{n_\text{layers}}\cdot(\frac{7k}{6}+\frac{1}{2})$, or a final overhead of 1.49x  which is in line with our effective measured runtime overhead of 1.51x (see Table~\ref{tab:results}).}

\section{DPR-style retriever training details}
\label{app:dpr_details}
\new{We followed the training recipe of DPR \cite{dpr} in training a BERT-base retriever with contrastive loss. The DPR objective requires positive and hard negatives to converge successfully, and here we use the top-1 scoring BM25 chunk as the positive example and the chunk ranked 5th by BM25 as the hard negative example. To ensure a fair comparison, we train our contrastive retriever on 16x more examples than the original DPR recipe describes.
}

\end{document}